%% file: acl_latex.tex
\pdfoutput=1

\documentclass[11pt]{article}

\usepackage[preprint]{acl}

\usepackage{times}
\usepackage{latexsym}

\usepackage[T1]{fontenc}

\usepackage[utf8]{inputenc}

\usepackage{microtype}

\usepackage{inconsolata}

\usepackage{graphicx}

\usepackage{hyperref}
\usepackage{url}
\usepackage{booktabs}
\usepackage{multirow}
\usepackage{amsmath}
\usepackage{amssymb}
\usepackage{mathtools}
\usepackage{amsthm}
\usepackage{blindtext}
\usepackage{tabularray}
\usepackage{longtable}
\usepackage{xltabular}
\usepackage{enumitem}
\usepackage{subcaption}
\usepackage[most]{tcolorbox}
\usepackage{float}
\usepackage{caption}
\usepackage{xcolor}

\usepackage{lineno}

\definecolor{darkblue}{rgb}{0, 0, 0.5}
\definecolor{pastelyellow}{HTML}{FFF5D6}
\definecolor{pastelgreen}{HTML}{D6EAC7}
\definecolor{pastelblue}{HTML}{D3E1FA}
\hypersetup{colorlinks=true, citecolor=darkblue, linkcolor=darkblue, urlcolor=darkblue}

\title{VerifiAgent: a Unified Verification Agent in Language Model Reasoning}

\author{
  Jiuzhou Han$^{{\natural} }$\ \ \ \ \ 
  Wray Buntine$^{{\flat}}$\ \ \ \ \ 
  Ehsan Shareghi$^{{\natural} }$\\
  $^{{\natural} }$~Department of Data Science \& AI, Monash University \\
  $^{{\flat}}$~College of Engineering and Computer Science, VinUniversity\\
  {jiuzhou.han@monash.edu}\ \ \ \ \ \ {wray.b@vinuni.edu.vn} \\{ehsan.shareghi@monash.edu}
}

\begin{document}
\maketitle
\begin{abstract}
Large language models demonstrate remarkable reasoning capabilities but often produce unreliable or incorrect responses. Existing verification methods are typically model-specific or domain-restricted, requiring significant computational resources and lacking scalability across diverse reasoning tasks. To address these limitations, we propose VerifiAgent, a unified verification agent that integrates two levels of verification: meta-verification, which assesses completeness and consistency in model responses, and tool-based adaptive verification, where VerifiAgent autonomously selects appropriate verification tools based on the reasoning type, including mathematical, logical, or commonsense reasoning. This adaptive approach ensures both efficiency and robustness across different verification scenarios. Experimental results show that VerifiAgent outperforms baseline verification methods (e.g., deductive verifier, backward verifier) among all reasoning tasks. Additionally, it can further enhance reasoning accuracy by leveraging feedback from verification results. VerifiAgent can also be effectively applied to inference scaling, achieving better results with fewer generated samples and costs compared to existing process reward models in the mathematical reasoning domain.\footnote{Code is available at \url{https://github.com/Jiuzhouh/VerifiAgent}.}
\end{abstract}

\section{Introduction}
\label{sec:intro}
\input{Intro}

\section{Related Work}
\label{sec:related_work}
\input{Related_Work}

\section{VerfiAgent}
\label{sec:verifiagent}
\input{Verifiagent}

\section{Experiment}
\label{sec:experiment}
\input{Experiment}

\section{Analysis}
\label{sec:analysis}

\input{Analysis}

\section{Conclusion}
\label{sec:conclusion}
\input{Conclu}

\section*{Limitations}

Our study has two primary limitations. First, due to the significant computational costs associated with accessing and running LLMs, we were unable to evaluate a comprehensive range of models, necessitating a selection of representative models. Second, the current implementation of VerifiAgent only supports three tools. While these demonstrate its core capabilities, we plan to expand the toolset in future work to enhance its versatility and applicability across more verification scenarios.



\bibliography{custom}

\appendix
\section*{Appendix}
\label{sec:Appendix}
\input{Appendix}

\end{document}

%% file: Intro.tex
Large language models (LLMs) have demonstrated significant capabilities in natural language reasoning tasks, exhibiting potential to solve complex problems across diverse domains~\citep{DBLP:journals/corr/abs-2412-15115, DBLP:journals/corr/abs-2501-12948, DBLP:journals/corr/abs-2407-21783, DBLP:journals/corr/abs-2303-08774}. However, despite their advanced reasoning abilities, these models often produce responses that are unreliable or incorrect, which poses substantial challenges for practical applications that require high precision and trustworthiness~\citep{DBLP:journals/natmi/AugensteinBCCCCDFHHHJMM24, DBLP:journals/air/HuangRHJDWBMQZCZWXWFM24}. To address this critical issue, several verification methods have been proposed, ranging from task-specific verifiers to generalised verification methods leveraging prompting techniques. 

Training a task-specific verifier to verify the output of LLM requires specific training data. For instance, \citet{DBLP:conf/icml/Ni0RSYWL23} train a verification model that judges language-to-code outputs based on both program text and execution results.
\citet{DBLP:journals/corr/abs-2410-05318} design verifiers trained on outputs from multiple reasoning paradigms, using correctness signals for improvement. \citet{DBLP:conf/acl/HanCBS24} train a lightweight verifier for improving semantic graph generation in text-to-graph tasks. Similarly, \citet{DBLP:journals/corr/abs-2409-16461} train a verifier that corrects potential syntactic and semantic first-order logic translation errors. Nevertheless, these existing solutions typically face limitations such as domain restriction, computational inefficiency, and lack of scalability when handling varied reasoning tasks. Table~\ref{tab:verifier_stat} demonstrates a feature comparison of various verification methods.

In this paper, we propose VerifiAgent, a unified verification agent designed specifically to overcome these limitations by offering a generalisable and efficient verification framework. Unlike prior methods, VerifiAgent adopts a two-layer verification mechanism, comprising meta-verification and tool-based adaptive verification. The meta-verification layer ensures completeness and logical consistency of responses, while the tool-based adaptive verification autonomously selects appropriate external tools (e.g., Python interpreters, symbolic solvers, search engines) to deal with different reasoning types, including mathematical, logical, commonsense, and hybrid reasoning tasks.

Our approach not only achieves superior verification accuracy compared to existing baseline methods, such as deductive verifier~\citep{DBLP:conf/nips/LingFLHLMS23}, backward verifier~\citep{DBLP:conf/emnlp/WengZX0HLSLZ23}, but also enhances reasoning accuracy by integrating detailed feedback derived from the verification process. Furthermore, VerifiAgent can be effectively applied to inference scaling, requiring significantly fewer computational resources compared to standard Process Reward Models (PRMs), thereby providing a practical approach to improve LLM performance during inference. Through extensive experiments across three types of reasoning tasks, we summarise two key empirical findings: 1) An LLM reasoner can improve via inference scaling methods like Majority Vote, PRMs, or VerifiAgent, but VerifiAgent achieves higher accuracy at lower cost. 2) VerifiAgent’s capabilities scale alongside improvements in its backbone LLM, enabling consistent performance gains on the same reasoner.

%% file: Related_Work.tex
\begin{table}[t]
\centering
\resizebox{0.48\textwidth}{!}{%
\begin{tabular}{@{}lcccc@{}}
\toprule
\textbf{Method}                                                          & \textbf{Training-free} & \textbf{Generalised} & \textbf{Tool-based} & \textbf{Fine-grained Feedback} \\ \midrule
DIVERSE \citep{DBLP:conf/acl/LiLZFCLC23}                 & $\times$               & $\times$             & $\times$            & $\times$                       \\
PiVe \citep{DBLP:conf/acl/HanCBS24}                      & $\times$               & $\times$             & $\times$            & $\checkmark$                   \\
Math/Code-Rev \citep{DBLP:journals/corr/abs-2410-05318} & $\times$               & $\times$             & $\checkmark$        & $\times$                       \\
LEVER \citep{DBLP:conf/icml/Ni0RSYWL23}                  & $\times$               & $\times$             & $\checkmark$        & $\times$                       \\
CoVe \citep{DBLP:conf/acl/DhuliawalaKXRLC24}             & $\checkmark$           & $\times$             & $\times$            & $\checkmark$                   \\
CSV\citep{DBLP:conf/iclr/ZhouWLSLQLJSZ024}              & $\checkmark$           & $\times$             & $\checkmark$        & $\checkmark$                   \\
Deductive Verifier \citep{DBLP:conf/nips/LingFLHLMS23}  & $\checkmark$           & $\checkmark$         & $\times$            & $\times$                       \\
Backward Verifier \citep{DBLP:conf/emnlp/WengZX0HLSLZ23} & $\checkmark$           & $\checkmark$         & $\times$            & $\times$                       \\
VerifiAgent (ours)                                                       & $\checkmark$           & $\checkmark$         & $\checkmark$        & $\checkmark$                   \\ \bottomrule
\end{tabular}}
\caption{Comparison of various verification methods.}
\label{tab:verifier_stat}
\end{table}

\subsection{LLMs as Verifiers}

Leveraging the prompting and in-contenxt learning ability of LLMs to verify the outputs of LLMs provides a generalised approach of verification. \citet{DBLP:journals/corr/abs-2405-14092, DBLP:conf/emnlp/WengZX0HLSLZ23} show that LLMs can refine reasoning chains via backward verification or masked condition checking, while \citet{DBLP:conf/nips/LingFLHLMS23} decompose solutions into verifiable steps using a Natural Program format. \citet{DBLP:conf/acl/DhuliawalaKXRLC24} propose Chain-of-Verification, which decomposes the verification into a sequence of questions, improving factual consistency through multi-step prompting. \citet{DBLP:conf/naacl/Hong0P0Z24} evaluate LLMs’ ability to detect logical errors, finding that while models can catch some flaws, their verification is often shallow. \citet{DBLP:journals/corr/abs-2402-08115} further investigate the reliability of self-critique, demonstrating that performance often degrades when doing self-verification without external grounding. 

To enhance the verification quality of LLMs, some methods integrate external tools. \citet{DBLP:conf/iclr/ZhouWLSLQLJSZ024} introduce code-based self-verification, which prompts GPT-4 Code Interpreter to evaluate and fix its answers by executing code and interpreting the output. Similarly, \citet{DBLP:conf/iclr/GouSGSYDC24} propose CRITIC, a framework where LLMs interact with tools (e.g., calculators, search engines) to critique and revise their own outputs, leading to improved factuality and reasoning. Different from these works, our VerifiAgent provides a generalised verification agentic framework adaptable to diverse reasoning tasks with fine-grained feedback.  

A broader concept of verification by LLMs is referred to LLM-as-a-Judge, where LLMs are used as general evaluators for tasks like response scoring, pairwise comparison, and content moderation. As surveyed by \citet{DBLP:journals/corr/abs-2411-15594, DBLP:journals/corr/abs-2411-16594}, LLM-as-a-Judge systems perform holistic evaluations, as a scalable and consistent alternative to human evaluation, applicable in model benchmarking, safety assessment, and alignment data labelling. VerifiAgent can be viewed as a specialised type of LLM-as-a-Judge system, specifically designed to evaluate the correctness of certain reasoning tasks through tool-based verification mechanisms.

\subsection{Scaling Test-Time Compute}
Scaling test-time compute refers to allocating more computational budget during inference via sampling, deeper reasoning, or adaptive search to boost model accuracy. \citet{DBLP:journals/corr/abs-2407-21787} present a comprehensive study of inference-time scaling through repeated sampling, demonstrating that coverage—the probability of generating at least one correct answer—scales log-linearly with the number of samples. \citet{DBLP:journals/corr/abs-2411-17501} theoretically analyse the limits of resampling, showing that imperfect verifiers lead to diminishing returns, especially when false positives dominate. Similarly, \citet{DBLP:journals/corr/abs-2502-12118} argue that verifier-based strategies scale more robustly than verifier-free ones, particularly when base models exhibit anti-concentrated output distributions.

The growing use of Process Reward Models (PRMs) suggests that fine-grained supervision over intermediate reasoning steps can improve model reliability~\citep{DBLP:conf/iclr/LightmanKBEBLLS24, DBLP:conf/acl/WangLSXDLCWS24, DBLP:journals/corr/abs-2501-07301}. PRMs offer another strategy for scaling test-time compute by enhancing Best-of-N sampling~\citep{DBLP:journals/corr/abs-2408-03314}. As a verifier, VerifiAgent does not require any training process, eliminating the need for collecting task-specific training data. By leveraging frozen LLMs, it can be integrated into test-time compute scaling strategies, enhancing the accuracy of LLM outputs.

%% file: Verifiagent.tex
\begin{figure*}[t]
\centering
\vspace{-10mm}
\includegraphics[width=\textwidth]{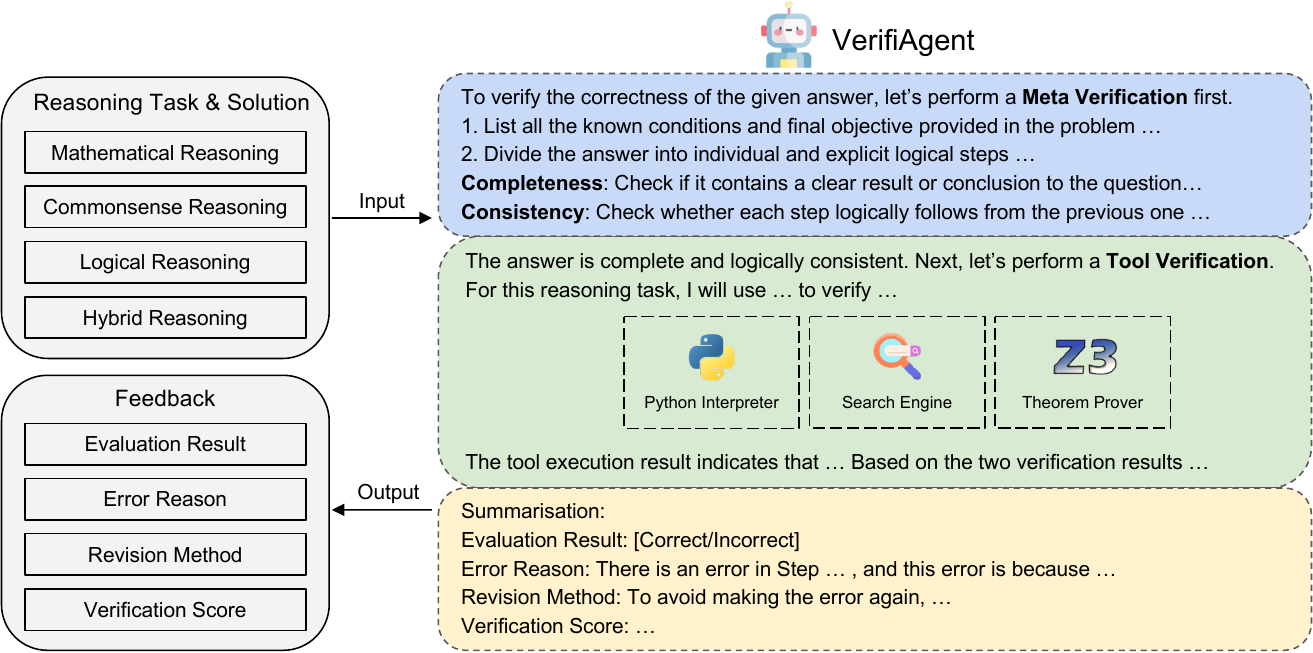}
\caption{An overview of VerifiAgent. Given a reasoning task and a candidate solution, VerifiAgent leverages two levels of verification: (1) meta verification -- verifying the completeness and consistency of the solution and (2) tool-based adaptive verification -- autonomously selecting appropriate tools to do the correctness verification. The VerifiAgent can provide fine-grained feedback about the verification process based on the instruction in the prompt.}
\label{fig:pipeline}
\vspace{-3mm}
\end{figure*}

VerifiAgent is a plug-and-play verification framework that empowers frozen LLMs to utilise external mechanisms to verify the correctness of solutions of diverse reasoning tasks. As illustrated in Figure~\ref{fig:pipeline}, VerifiAgent adopts a two-layer verification mechanism, which contains two levels of verification. The first level is to do a Meta Verification, and the second level is to do a Tool-based Adaptive Verification. The solution will be evaluated sequentially through the two layers. The second-level verification can further validate the results from meta-verification, enhancing the accuracy of the verification results. With this two-layer verification mechanism, the VerifiAgent can provide fine-grained feedback of the verification process.

\subsection{Meta Verification}
The Meta Verification aims to verify two aspects of the solution: completeness and consistency. The completeness refers to a solution that is self-contained, fully addresses every part of the question, and contains a clear result or conclusion. The consistency refers to reasoning that follows a logical structure with no jumps, gaps, or inconsistencies. This initial layer acts as a foundational check, preventing incomplete or inconsistent solutions from progressing further. Through meta verification, VerifiAgent ensures that only solutions with structural integrity and coherent reasoning proceed to the next tool-based adaptive verification.

Since the solutions of different types of reasoning tasks may have different structures, to make VerifiAgent adaptable to diverse solutions, we leverage a unified way to rewrite the solutions in the meta verification phase. Specifically, the agent will first list all the known conditions and the final objective provided in the problem, then divide the solution into individual and explicit logical steps. This will be beneficial for the meta verification and the following tool-based adaptive verification. See Appendix~\ref{appd:examples} for examples.

\subsection{Tool-based Adaptive Verification}
After the meta verification stage, the solution enters the Tool-based Adaptive Verification phase. This level leverages external tools, such as Python program interpreter, search engine and symbolic solver, to cross-check the correctness of the solution. The agent will first solve the question using appropriate tools, and then verify the results by comparing them with the original solution. Unlike the meta verification stage, which evaluates general reasoning quality, this phase evaluates factual and computational accuracy.

VerifiAgent dynamically selects the most suitable verification tool based on the nature of the task and instructions. For instance, in mathematical reasoning, it may utilise a Python interpreter to verify calculations, while for knowledge-based commonsense reasoning, it may query a search engine to gather relevant information. For hybrid reasoning tasks, it can combine multiple tools to ensure comprehensive verification. Additionally, VerifiAgent autonomously determines the required number of external tool calls, continuing until it gathers sufficient information to validate the answer.

When VerifiAgent selects a tool for verification, the environment returns the corresponding execution result. Based on this observation, the agent iteratively determines its next action until the verification process is complete. The VerifiAgent not only ensures the accuracy of solutions but also provides a transparent and interpretable verification process for natural language reasoning tasks. See Appendix~\ref{appd:examples} for examples.

\subsection{Fine-grained Feedback}
Based on the two levels of verification, VerifiAgent provides a final evaluation result (i.e., $Correct/Incorrect$) to indicate the correctness of the solution. In addition to the verification result, VerifiAgent also generate a $V_{score}$ as a confidence score of the verification. $V_{score}$ is calculated by applying the softmax function to the log probability of the token $(Correct/Incorrect)$ and the log probabilities of the top 5 alternative tokens. Specifically:
$$
V_{score}=\frac{\exp \left(p\left(t\right)\right)}{\sum_{k=1}^5 \exp \left(p\left(t_k\right)\right)}
$$
where $V_{score}$ represents the confidence score for the verification result token $t$. the term $p\left(t\right)$ denotes the log probability of the token $t$ generated by the LLM. $p\left(t_k\right)$ is for $k=1$ to $5$ represents the log probabilities of the top five predicted tokens at the verification result token position. This equation ensures that the confidence score reflects the relative likelihood of the chosen token compared to the top alternatives, effectively normalising the scores within the range of 0 to 1.

Furthermore, when a solution is deemed incorrect, VerifiAgent provides fine-grained feedback about the verification process based on the instruction in the prompt. This feedback includes the identified error reason, derived from the two levels of verification, and a potential revision method that incorporates observations from tool execution results (See Appendix~\ref{appd:examples} for examples). Such feedback can be leveraged to refine and enhance the solution, improving the accuracy of reasoning tasks.

%% file: Experiment.tex
\subsection{Baseline and Experimental Setup}

\noindent\textbf{Datasets.} We evaluate VerifiAgent on three natural language reasoning tasks, including mathematical reasoning, logical reasoning, commonsense reasoning, and hybrid reasoning. Specifically, for mathematical reasoning we use GSM8K~\citep{DBLP:journals/corr/abs-2110-14168} and MATH~\citep{DBLP:conf/nips/HendrycksBKABTS21}, for logical reasoning we use FOLIO~\citep{DBLP:conf/emnlp/HanS0QRZCPQBSWS24} and ProverQA~\citep{Qi2025LargeLM}, for commonsense reasoning we use HotpotQA~\citep{DBLP:conf/emnlp/Yang0ZBCSM18} and StrategyQA~\citep{DBLP:journals/tacl/GevaKSKRB21}, for hybrid reasoning we use ReWild~\citep{DBLP:conf/emnlp/YangXPSF24}. The statistics of the datasets are shown in Appendix~\ref{appd:data_stat}.

\noindent\textbf{Baselines.} Since VerifiAgent is a training-free and generalised approach, we compare it against baseline methods that are similarly prompting-based and generalised. Specifically:

\begin{itemize}[nosep, left=0pt, labelsep=1em]
  \item \textbf{Vanilla Verifier.} Vanilla Verifier employs a structured prompt to instruct the LLM to verify a solution given a problem, without relying on specialised mechanisms~\citep{DBLP:journals/tacl/KamoiZZHZ24}. 
  \item \textbf{Deductive Verifier.} Deductive Verifier~\citep{DBLP:conf/nips/LingFLHLMS23} enables the LLM to carry out explicit and rigorous deductive reasoning to evaluate the correctness of a solution. It decomposes the verification process into a sequence of step-by-step subprocesses using Natural Program, a natural language-based deductive reasoning format, to facilitate the breakdown of logical steps in a step-by-step manner. 
  \item \textbf{Backward Verifier.} Backward Verifier~\citep{DBLP:conf/emnlp/WengZX0HLSLZ23} appends the predicted answer to the question while masking the original condition, then prompts the LLM to predict the masked condition. Verification is conducted by comparing the predicted condition with the original one. If the two conditions align, the solution is deemed correct; otherwise, inconsistencies indicate errors in the provided solution.
\end{itemize}

\noindent\textbf{Models.} We explored various combinations of backbone LLMs for both the Reasoner and VerifiAgent. For the Reasoner, we utilise GPT-4o, o3-mini, and Llama-3.3-70B-Instruct-Turbo, while for the VerifiAgent, we employ GPT-4o and o1-mini. In our experiments, unless explicitly stated otherwise, both the Reasoner and VerifiAgent default to GPT-4o as their backbone LLMs.

\subsection{Main Result}

 \begin{table*}[t]\small
\centering
\vspace{-8mm}
\resizebox{\textwidth}{!}{%
\begin{tabular}{lccccccccccccc}
\toprule
\multirow{3}{*}{\textbf{Type}}         & \multirow{3}{*}{\textbf{Dataset}} & \multicolumn{9}{c}{\textbf{Baselines}}                                                                                         & \multicolumn{3}{c}{\multirow{2}{*}{\textbf{VerifiAgent}}} \\ \cmidrule{3-11}
                              &                          & \multicolumn{3}{c}{\textbf{Vanilla Verifier}} & \multicolumn{3}{c}{\textbf{Deductive Verifier}} & \multicolumn{3}{c}{\textbf{Backward Verifier}} & \multicolumn{3}{c}{}                             \\ \cmidrule{3-14} 
                              &                          & Acc        & Pre        & Rec        & Acc         & Pre         & Rec        & Acc         & Pre        & Rec        & Acc            & Pre            & Rec            \\ 
\midrule
\multirow{2}{*}{Mathematical} & GSM8K                    & 0.93       & 0.96       & 0.96       & 0.95        & 0.96        & 0.99       & 0.95        & 0.96      & 0.98       & \textbf{0.96}  & \textbf{0.96}  & \textbf{1.00}  \\
                              & MATH                        & 0.75       & 0.73       & 0.86       & 0.80        & 0.76        & 0.86       & 0.82        & 0.80      &0.88       & \textbf{0.85}  & \textbf{0.86}  & \textbf{0.92}  \\
\midrule
\multirow{2}{*}{Logical}      & FOLIO                              & 0.75       & 0.78       & 0.96       & 0.73        & 0.73        & 0.95       & 0.74        & 0.76      & 0.96       & \textbf{0.76}  & \textbf{0.78}  & \textbf{0.97}  \\
                              & ProverQA                            & 0.75       & 0.77       & 0.97       & 0.74        & 0.75        & \textbf{0.98}       & 0.75        & 0.78      & 0.96       & \textbf{0.77}  & \textbf{0.82}  & 0.95  \\
\midrule
\multirow{2}{*}{Commonsense}  & StrategyQA                              & 0.78       & 0.79       & 0.92       & 0.75        & 0.82        & 0.92       & 0.79        & 0.80      & 0.94       & \textbf{0.84}  & \textbf{0.85}  & \textbf{0.95}  \\
                              & HotpotQA                               & 0.56       & 0.53       & 0.91       & 0.56        & 0.53        & \textbf{0.96}       & 0.57        & 0.54      & 0.90       & \textbf{0.61}  & \textbf{0.56}  & 0.92  \\
\midrule
Hybrid                        & ReWild                                  & 0.76       & 0.88       & 0.82       & 0.61        & \textbf{0.91}        & 0.60       & 0.74        & 0.87      &0.84       & \textbf{0.78}  & 0.88  & \textbf{0.89}  \\
\bottomrule
\end{tabular}}
\caption{Main results of VerifiAgent on different reasoning tasks. The evaluation metrics are accuracy (Acc), precision (Pre), and recall (Rec). \textbf{Bold} shows the best result for each row.}
\label{tab:main}
\end{table*}

Table~\ref{tab:main} shows the performance of VerifiAgent compared to baseline methods (Vanilla, Deductive, and Backward Verifiers) across different reasoning tasks. Overall, VerifiAgent consistently outperforms baselines, excelling in accuracy while maintaining competitive precision and recall across mathematical, logical, commonsense, and hybrid reasoning tasks. Specifically, for mathematical reasoning tasks, VerifiAgent attains the highest accuracy (0.96 and 0.85) and recall scores (1.00 and 0.92) on GSM8K and MATH datasets, respectively. In logical reasoning, VerifiAgent demonstrates improvements, particularly on FOLIO (accuracy 0.76, recall 0.97) and ProverQA (precision 0.82). For commonsense reasoning tasks, VerifiAgent significantly outperforms baselines on StrategyQA with accuracy and precision of 0.84 and 0.85, respectively, while remaining competitive on HotpotQA. Finally, on the hybrid reasoning dataset ReWild, VerifiAgent achieves the best accuracy (0.78) and recall (0.89), highlighting its verification capabilities in handling complex reasoning tasks. To investigate the impact of different backbone LLMs on VerifiAgent’s performance, we further evaluate VerifiAgent using o1-mini as an alternative backbone model. The results indicate that the verification capability of VerifiAgent scales effectively with the underlying backbone model’s capacity. Due to the page limit, we put the detailed results in Appendix~\ref{appd:diff_backbone_llms}.

\begin{table*}[t]
\centering
\begin{tabular}{@{}lccc@{}}
\toprule
\textbf{Method}                                & \textbf{MATH} & \textbf{ProverQA} & \textbf{StrategyQA} \\ \midrule
GPT-4o Reasoner                                & 69.4(1)      & 75.3(1)          & 84.2(1)            \\
- IS w/ Majority Vote @10       & 73.5(10)     & 77.0(10)          & 85.6(10)           \\
- IS w/ VerifiAgent (GPT-4o)  & 74.0(1.5)    & 77.3(1.6)        &   86.0(1.3)          \\
- IS w/ VerifiAgent (o1-mini) &    78.0(1.8)           &                  77.7(1.3) &          87.3(1.2)            \\ \midrule
o3-mini Reasoner                               &      87.9(1)         &   78.3(1)               &        76.4(1)            \\
- IS w/ Majority Vote @8       &      91.1(10)         &        80.0(10)           &         78.2(10)            \\
- IS w/ VerifiAgent (GPT-4o)  &        88.3(1.3)       &                79.1(1.1)   &             78.6(1.3)        \\
- IS w/ VerifiAgent (o1-mini) &       91.4(1.1)        &                80.7(1.1)   &        79.0(1.6)             \\ \midrule
Llama-3.3-70B-Instruct-Turbo Reasoner                            &     62.3(1)          &    70.6(1)               &            83.8(1)         \\
- IS w/ with Majority Vote @10       &        68.3(10)       &    71.7(10)              &         84.7(10)            \\
- IS w/ VerifiAgent (GPT-4o)  &        69.7(2.0)       &       72.0(1.3)            &         85.1(1.3)            \\
- IS w/ VerifiAgent (o1-mini) &        71.1(2.2)       &       74.0(1.3)             &    85.1(1.4)                 \\ \bottomrule
\end{tabular}
\caption{Results of different Reasoners with Inference Scaling (IS) methods on three datasets. The number in the bracket denotes the average number of samples for each question.}
\label{tab:inf_scal}
\end{table*}

\subsection{Inference Scaling with VerifiAgent}
Inference scaling aims at enhancing reasoning performance by utilising increased computational resources during the inference stage. However, this approach inherently requires effective verification to ensure the accuracy and reliability of generated answers~\citep{Setlur2025ScalingTC}. Due to the verification ability of VerifiAgent, it naturally complements inference scaling approaches by serving as an effective verifier during the inference process. Specifically, we first sample an output from the LLM. If this output passes verification by the VerifiAgent, the process terminates; otherwise, we continue sampling additional candidate outputs until one passes verification or the maximum number of samples is reached. For cases reaching the maximum number of samples, we select the final answer using a majority vote approach. We compare our VerifiAgent-based inference scaling method with the standard Majority Vote approach that does not employ a verifier. Majority Vote aggregates multiple sampled responses directly from reasoners without any verification. Table~\ref{tab:inf_scal} demonstrates the performance across three reasoning datasets (MATH, ProverQA, and StrategyQA) using various combinations of reasoners (GPT-4o, o3-mini, and Llama-3.3-70B-Instruct-Turbo) and VerifiAgent variants (GPT-4o and o1-mini).

Across all datasets and reasoners, inference scaling with VerifiAgent consistently outperforms Majority Voting, achieving higher accuracy with fewer samples and less cost (See Appendix~\ref{appd:infer_cost}). Notably, o3-mini reasoner achieves the highest performance on MATH and ProverQA but the lowest on StrategyQA among all reasoners, suggesting that o3-mini is more proficient in mathematical and logical reasoning than in knowledge-intensive commonsense reasoning tasks. We identify two key findings: (1) When the reasoner and VerifiAgent are the same model (e.g., GPT-4o) or have comparable capacities (e.g., Llama-3.3-70B-Instruct-Turbo paired with GPT-4o), integrating VerifiAgent significantly enhances performance, with further improvement achievable by employing a stronger VerifiAgent (e.g., o1-mini). (2) When the reasoner (o3-mini) surpasses the VerifiAgent (GPT-4o) in capability, the performance gain is limited. However, pairing a strong reasoner with a stronger VerifiAgent (o1-mini) substantially enhances performance.

\begin{figure}[t]
    \centering
    \begin{subfigure}[h]{0.479\textwidth}
        \centering
        \includegraphics[width=\textwidth]{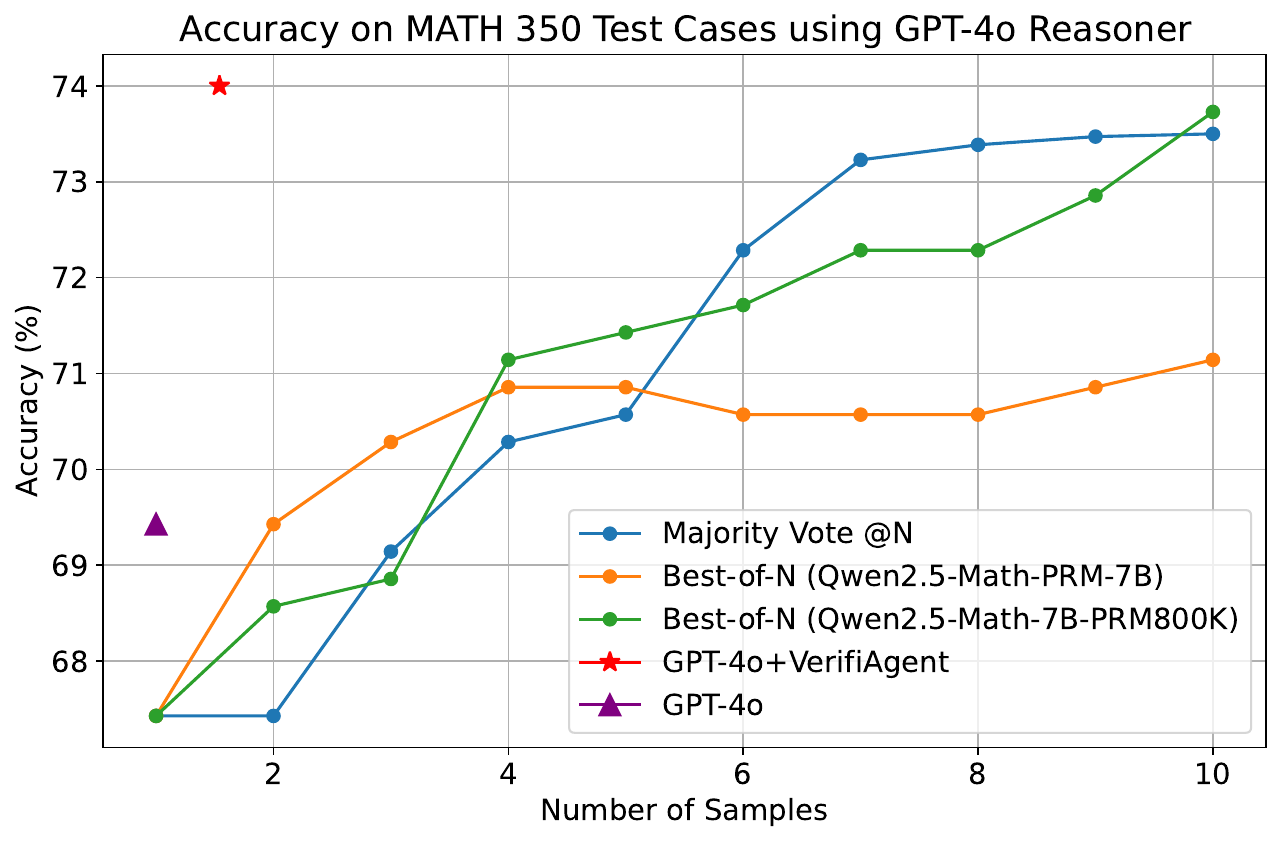}
        \caption{GPT-4o Reasoner}
        \label{fig:gpt4o_prm_plot}
    \end{subfigure}
    \hfill
    \begin{subfigure}[h]{0.515\textwidth}
        \centering
        \includegraphics[width=\textwidth]{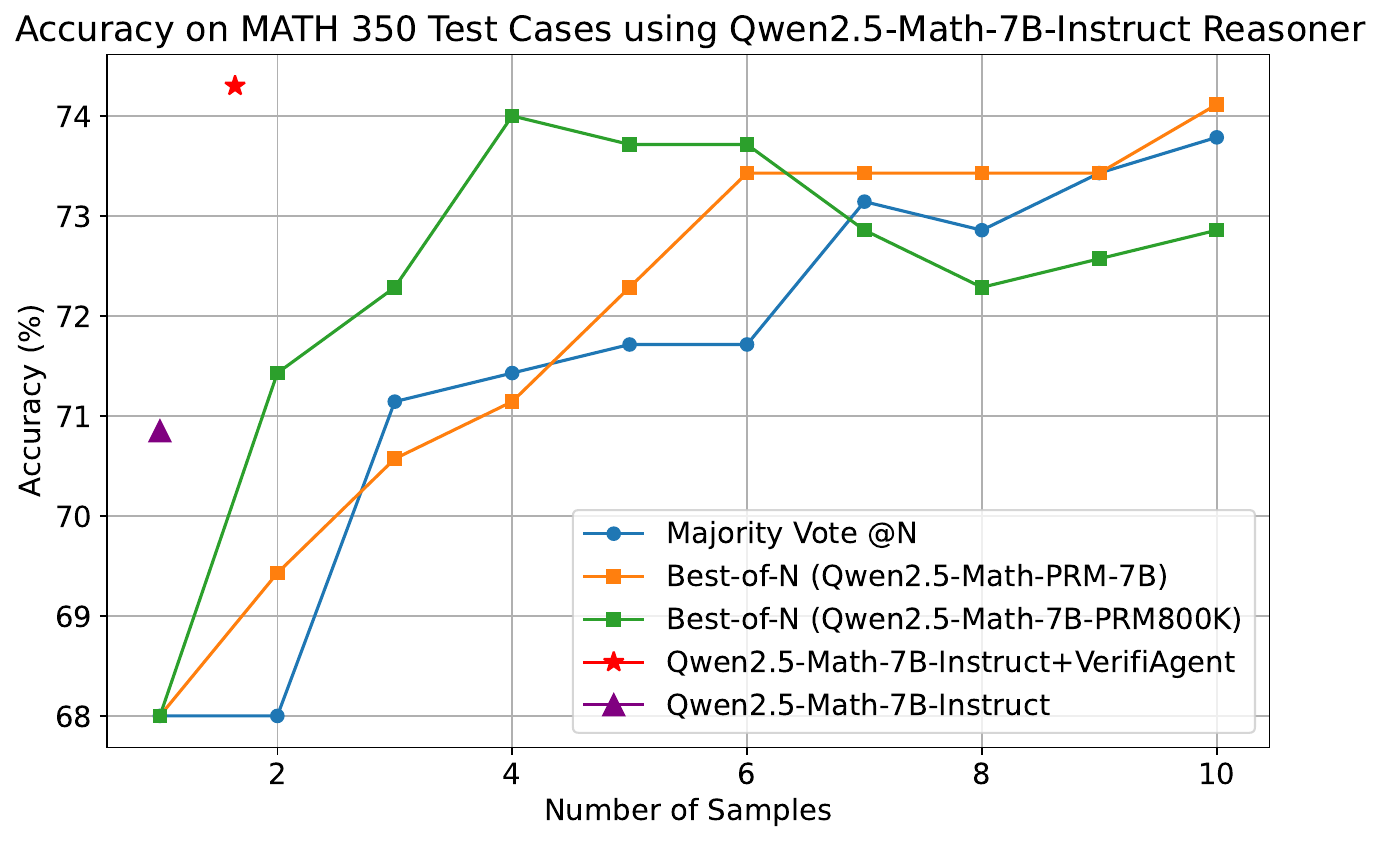}
        \caption{Qwen2.5-Math-7B-Instruct Reasoner}
        \label{fig:qwen_prm_plot}
    \end{subfigure}
    \caption{Results of GPT-4o Reasoner and Qwen2.5-Math-7B-Instruct Reasoner with different inference scaling methods on MATH. VerifiAgent uses GPT-4o as the backbone LLM.}
    \vspace{-3mm}
    \label{fig:prm_plots}
\end{figure}

PRMs provide another approach to inference scaling. We investigated two open-source PRMs specifically designed for the MATH dataset: Qwen2.5-Math-PRM-7B and Qwen2.5-Math-7B-PRM800K, which fine-tune Qwen2.5-Math-7B-Instruct using synthetic data from Qwen models~\citep{DBLP:journals/corr/abs-2501-07301} and PRM800K~\citep{DBLP:conf/iclr/LightmanKBEBLLS24}, respectively. These PRMs assign scores to each reasoning step, and we use the last step score as the final response score. Following previous studies ~\citep{DBLP:journals/corr/abs-2501-07301, DBLP:conf/iclr/LightmanKBEBLLS24, DBLP:journals/corr/abs-2409-12122, DBLP:conf/acl/WangLSXDLCWS24}, we evaluate the PRMs using the Best-of-N sampling strategy, selecting the highest-scored response from N candidates according to a PRM. The evaluation results for the GPT-4o and Qwen2.5-Math-7B-Instruct reasoners are shown in Figure~\ref{fig:prm_plots}. As the number of samples increases, both Majority Vote and Best-of-N sampling strategies consistently improve in accuracy. When sampling 10 responses, the Best-of-N method's accuracy approaches that of VerifiAgent, which notably achieves comparable performance with significantly fewer average samples (1.5 and 1.6 on GPT-4o reasoner and Qwen2.5-Math-7B-Instruct reasoner, respectively).

Interestingly, the two PRMs exhibit distinct behaviours depending on the reasoner. For the GPT-4o reasoner, Qwen2.5-Math-7B-PRM800K significantly outperforms Qwen2.5-Math-PRM-7B, which even underperforms relative to the Majority Vote baseline. However, for the Qwen2.5-Math-7B-Instruct reasoner, Qwen2.5-Math-PRM-7B outperforms Qwen2.5-Math-7B-PRM800K at 10 samples. These results indicate that GPT-4o benefits more from Qwen2.5-Math-7B-PRM800K, whereas Qwen2.5-Math-7B-Instruct gains greater improvements from Qwen2.5-Math-PRM-7B. We hypothesise that the linguistic discrepancies may affect the performance of PRMs. Specifically, Qwen2.5-Math-7B-PRM800K utilises synthetic data from GPT-style LLMs for training, while Qwen2.5-Math-PRM-7B employs data generated by Qwen-style LLMs for training. 

\begin{table}[t]
\centering
\vspace{-2mm}
\resizebox{0.48\textwidth}{!}{%
\begin{tabular}{lccc}
\toprule
& \textbf{MATH} & \textbf{ProverQA} & \textbf{StrategyQA} \\ \midrule
Init. Reasoning Acc.& 69.4 & 75.3 & 84.3 \\ \midrule
\textbf{Feedback Type} &\multicolumn{3}{c}{\textbf{Precaution-Based Feedback}} \\ \midrule
Verification Result & 69.7 & 76.0 & 84.3 \\
+ Error Reason& 74.9 & 77.0 & 85.6 \\
+ Mitigation Method & 73.4 & 77.6 & 86.0 \\ \midrule
\textbf{Feedback Type} & \multicolumn{3}{c}{\textbf{Post-Editing-Based Feedback}} \\ \midrule
Verification Result & 71.7 & 77.3 & 84.7 \\
+ Error Reason & 72.3 & 74.7 & 84.3 \\
+ Mitigation Method & 72.6 & 74.3 & 83.8 \\ \bottomrule
\end{tabular}}
\caption{Results of feedback utilisation on GPT-4o.}
\label{tab:feedback_results}
\end{table}

\subsection{Exploration on Feedback Utilisation}
VerifiAgent provides fine-grained feedback during verification, which includes an explicit error reason and a suggested revision method for enhancing solutions. To evaluate the effectiveness of this feedback, we conducted experiments using two distinct methods: precaution-based and post-editing-based feedback. In the precaution-based method, the LLM leverages feedback from previous verification attempts to proactively generate a new solution. Conversely, the post-editing-based method allows the LLM to directly refine its previous incorrect solution based on feedback provided. 

We explored three feedback settings for each method: (1) verification result only (i.e., simply indicating ``Incorrect''), (2) verification result with error reason, and (3) verification result with both error reason and revision method. Experiments were conducted on instances initially identified as ``Incorrect'' by VerifiAgent, and the results are presented in Table~\ref{tab:feedback_results}.

Overall, precaution-based feedback consistently outperforms post-editing-based feedback, indicating the inherent difficulty for LLMs to effectively correct previously incorrect responses. Additionally, within precaution-based feedback, providing richer information typically yields greater improvements. For post-editing-based feedback, however, mathematical reasoning tasks benefit from more detailed feedback, whereas logical and commonsense reasoning tasks achieve better performance with simpler, less detailed feedback.

\subsection{VerifiAgent on Tool-based Reasoner}

\begin{table}[t]
\centering
\resizebox{0.48\textwidth}{!}{%
\begin{tabular}{@{}lccc@{}}
\toprule
\textbf{Method}           & \textbf{MATH} & \textbf{ProverQA} & \textbf{StrategyQA} \\ \midrule
GPT-4o (Tool-use)            & 56.9 & 47.8     & 83.5       \\
GPT-4o (Tool-use) + Feedback & 61.5 & 50.3     & 85.7       \\
GPT-4o (CoT)                 & 69.4 & 75.3     & 84.3       \\
GPT-4o (CoT) + Feedback      & 73.4 & 77.6     & 86.0       \\ \bottomrule
\end{tabular}}
\caption{Resutls of GPT-4o reasoner using CoT and Tool with feedback on three different reasoning tasks.}
\label{tab:tool_reasoner}
\end{table}

\begin{table}[t]
\centering
\resizebox{0.35\textwidth}{!}{%
\begin{tabular}{@{}lccc@{}}
\toprule
\textbf{Dataset (Method)} & \textbf{Acc} & \textbf{Pre} & \textbf{Rec} \\ \midrule
MATH (CoT)                & 0.85         & 0.86         & 0.92         \\
MATH (Tool-use)           & 0.82         & 0.83         & 0.90         \\ \midrule
ProverQA (CoT)            & 0.77         & 0.82         & 0.95         \\
ProverQA (Tool-use)       & 0.75         & 0.79         & 0.93         \\ \midrule
StrtegyQA (CoT)           & 0.84         & 0.85         & 0.95         \\
StrtegyQA (Tool-use)      & 0.80         & 0.84         & 0.92         \\ \bottomrule
\end{tabular}}
\caption{Results of VerifiAgent evaluating CoT and Tool-use outputs on different reasoning tasks.}
\label{tab:tool_reasoner_verify}
\end{table}

\begin{figure*}[t]
    \centering
    \vspace{-12mm}
    \begin{minipage}[t]{0.48\textwidth}
        \centering
        \captionsetup{width=\textwidth}
        \includegraphics[width=0.8\linewidth]{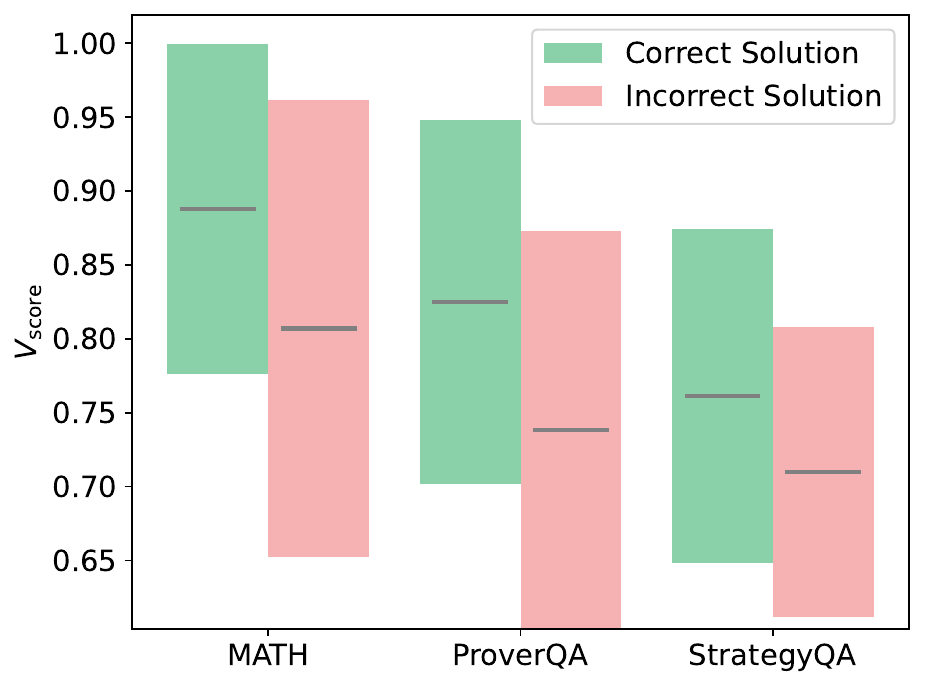}
        \caption{The floating bar chart comparing $V_{score}$  distributions (mean ± std) for correct and incorrect solutions across three datasets. The horizontal grey line indicates the mean.}
        \label{fig:vscore}
    \end{minipage}
    \hspace{1em}
    \begin{minipage}[t]{0.48\textwidth}
        \centering
        \captionsetup{width=\textwidth}
        \includegraphics[width=0.8\linewidth]{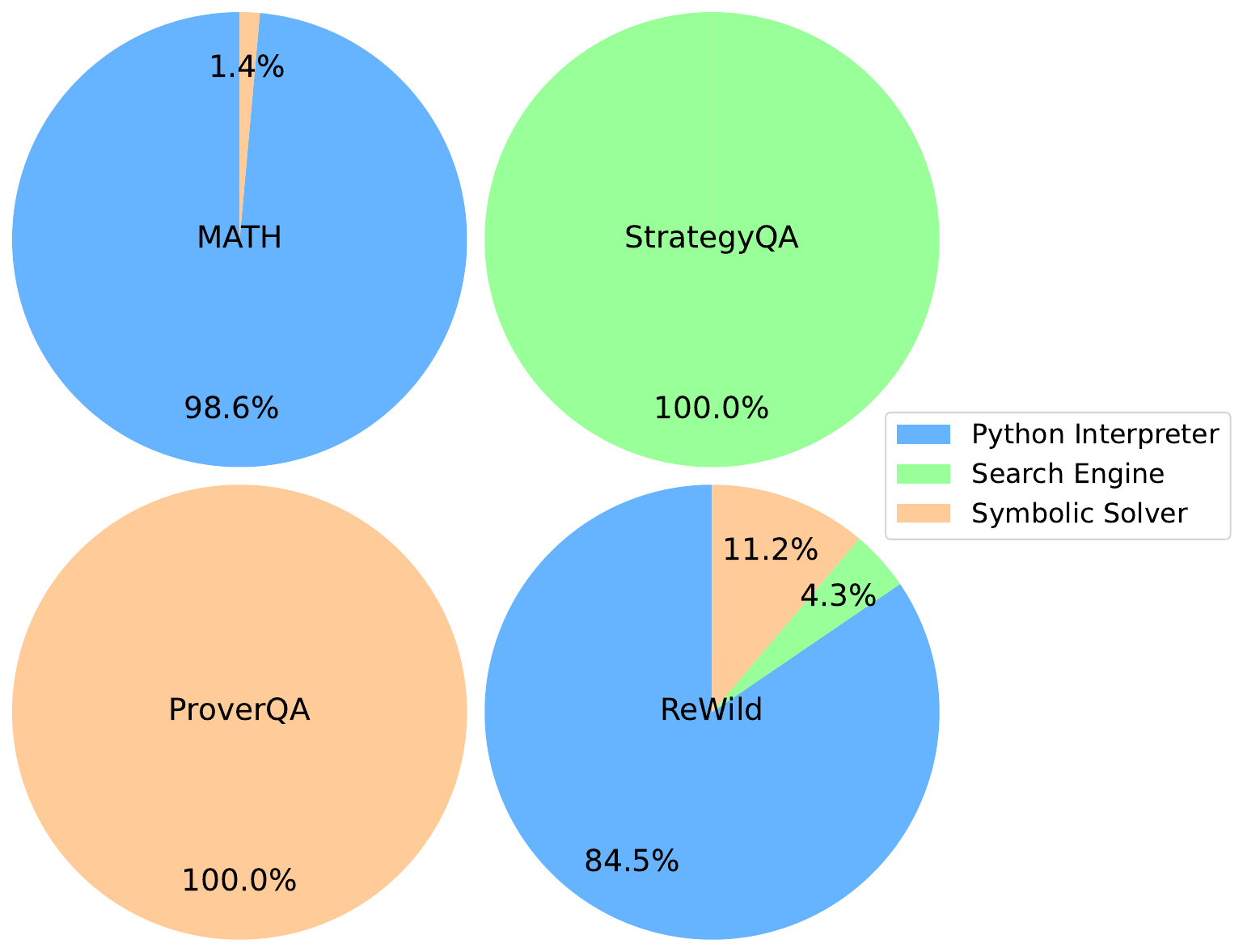}
        \caption{The pie charts showing the relative usage frequency of three different tools by the VerifiAgent across four types of reasoning tasks.}
        \label{fig:tool_stat}
    \end{minipage}
\end{figure*}

To evaluate the effectiveness of VerifiAgent on tool-based reasoning tasks, we conducted experiments using a tool-using reasoner with access to the same tools as VerifiAgent. Specifically, we evaluated GPT-4o on MATH, ProverQA, and StrategyQA, where the model was instructed to use the Python interpreter, Z3 Theorem Prover, and Search Engine, respectively. We then applied VerifiAgent to these outputs, leveraging its feedback to further improve performance. The results are shown in Table~\ref{tab:tool_reasoner}.

Interestingly, the tool-use baseline does not outperform CoT, especially in math and logic reasoning. This trend is also observed in prior works~\cite{DBLP:conf/iclr/YaoZYDSN023, DBLP:conf/acl/HanBS24}, where tool-augmented methods such as ReAct can underperform compared to CoT. We hypothesise this is due to several factors: (1) For simpler problems, LLMs may already solve them accurately via CoT, and mandatory tool use may introduce unnecessary complexity and more opportunities for errors. (2) For more difficult tasks, the LLM may still struggle to solve them effectively, even with tools. (3) The external knowledge from inaccurate tool-use can sometimes mislead the LLM’s correct prior knowledge. Despite these challenges, integrating VerifiAgent feedback consistently improves tool-use accuracy across all datasets, though results still lag behind CoT+Feedback in math and logic. For commonsense tasks (StrategyQA), tool-use approaches CoT+Feedback performance.

We also measured VerifiAgent’s verification performance on both CoT and tool-use outputs, shown in Table~\ref{tab:tool_reasoner_verify}. While performance is slightly lower for tool-use outputs, which is expected, since VerifiAgent is not designed specifically for program or tool-use evaluation. This points an important direction for future work, enhancing VerifiAgent’s robustness in verifying tool-based reasoning.

\subsection{Ablation Study}
Meta verification and tool verification are two essential components of VerifiAgent. To evaluate the individual contributions of these components, we conducted an ablation study, with results presented in Table~\ref{tab:ablation}. Results demonstrate that removing either meta verification or tool verification consistently reduces VerifiAgent's performance across all datasets. Specifically, omitting meta verification leads to noticeable declines in overall accuracy, while removing tool verification results in even more substantial performance reductions, bringing the performance close to baseline levels. Additionally, tool verification tends to enhance accuracy and precision, whereas meta verification primarily improves recall. These findings underscore the complementary roles of meta and tool verification, with each contributing uniquely to the effectiveness of VerifiAgent.

\begin{table}[t]
\centering
\vspace{-2mm}
\resizebox{0.48\textwidth}{!}{%
\begin{tabular}{@{}lccccccccc@{}}
\toprule
\multirow{2}{*}{\textbf{Method}} & \multicolumn{3}{c}{\textbf{MATH}} & \multicolumn{3}{c}{\textbf{ProverQA}} & \multicolumn{3}{c}{\textbf{StragegyQA}} \\ \cmidrule(l){2-10} 
                                 & \textbf{Acc} & \textbf{Pre} & \textbf{Rec} & \textbf{Acc} & \textbf{Pre} & \textbf{Rec} & \textbf{Acc} & \textbf{Pre} & \textbf{Rec} \\ \midrule
Vanilla Verifier                 & 0.75         & 0.73         & 0.86         & 0.75         & 0.77         & 0.97         & 0.78         & 0.79         & 0.92         \\
Deductive Verifier              & 0.80         & 0.76         & 0.86         & 0.74         & 0.75         & 0.98         & 0.75         & 0.82         & 0.92         \\
Backward Verifier               & 0.82         & 0.80         & 0.88         & 0.75         & 0.78         & 0.96         & 0.79         & 0.80         & 0.94         \\ \midrule
VerifiAgent                     & 0.85         & 0.86         & 0.92         & 0.77         & 0.82         & 0.95         & 0.84         & 0.85         & 0.95         \\
- w/o meta v.       & 0.79         & 0.78         & 0.96         & 0.74         & 0.81         & 0.90         & 0.83         & 0.83         & 0.94         \\
- w/o tool v.         & 0.75         & 0.75         & 0.98         & 0.74         & 0.75         & 0.98         & 0.78         & 0.80         & 0.96         \\ \bottomrule
\end{tabular}}
\caption{Ablation study results of VerifiAgent.}
\label{tab:ablation}
\vspace{-4mm}
\end{table}

%% file: Analysis.tex
\noindent\textbf{Verification Score Visualisation.} Figure~\ref{fig:vscore} visualises the $V_{score}$ for correct and incorrect solutions across three datasets. As illustrated, the mean $V_{score}$ for correct solutions is slightly higher than for incorrect ones on all the reasoning tasks. Since $V_{score}$ represents the confidence of the verification result, this indicates that the VerifiAgent is more confident when identifying correct solutions compared to incorrect ones. Additionally, the consistently lower variance in $V_{score}$ among correct solutions further supports the reliability of the agent in verifying correct responses.

\noindent\textbf{Tool Usage Analysis.} VerifiAgent autonomously determines the reasoning type of a task and selects the appropriate tool for verification. Figure~\ref{fig:tool_stat} illustrates tool usage across four reasoning task types. For the MATH dataset (mathematical), the Python Interpreter is predominantly used (98.6\%), with minimal reliance on the Symbolic Solver (1.4\%), reflecting the computational nature of the task. StrategyQA (commonsense) exclusively relies on the Search Engine (100\%), highlighting its dependence on external knowledge for the verification. ProverQA (logical) solely utilises the Symbolic Solver (100\%), aligning with its need for logical and symbolic reasoning. ReWild (hybrid) shows a more balanced tool distribution, primarily using the Python Interpreter (84.5\%), supplemented by the Symbolic Solver (11.2\%) and the Search Engine (4.3\%). The results demonstrate that the VerifiAgent effectively selects appropriate external tools based on the nature of the reasoning task.

\begin{figure*}[t]
\centering
\includegraphics[width=1.0\textwidth]{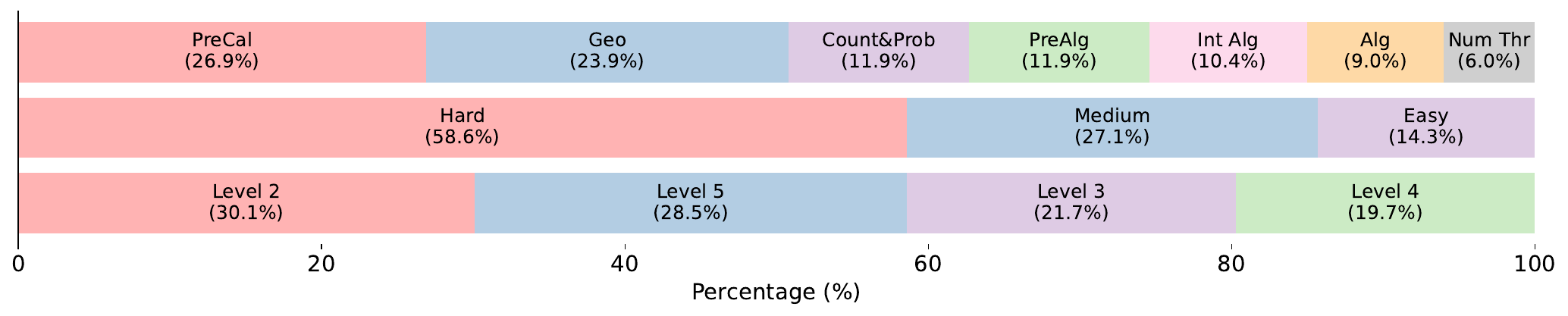}
\caption{The proportion of different question types among VerifiAgent’s incorrectly verified examples by GPT-4o Reasoner. From top to bottom, the bars represent MATH, ProverQA, and StrategyQA datasets, respectively. For MATH and ProverQA, the number of questions in each type is the same. For the imbalanced StrategyQA, the proportion is normalised by the total number of questions per difficulty level.}
\label{fig:error_stat}
\end{figure*}

\noindent\textbf{Error Analysis.} To further investigate the capability of VerifiAgent, we conducted an error analysis on different types of questions. The MATH dataset contains seven types of math problems: Algebra (Alg), Counting\&Probability (Count\&Prob), Geometry (Geo), Intermediate Algebra (Int Alg), Number Theory (Num Thr), Prealgebra (PreAlg) and Precalculus (PreCal). The ProverQA classifies the question into three types based on the difficulty level: Hard, Medium and Easy. Although StrategyQA does not explicitly label questions by difficulty, each question includes a decomposition into sub-questions that reflect its reasoning pathway. We used the number of decomposed sub-questions (ranging from 1 to 5) as an indicator of question difficulty, classifying them into five levels (Level 1 through Level 5).

Figure~\ref{fig:error_stat} illustrates the distribution of question types among cases where VerifiAgent provided incorrect verifications. In the MATH dataset, Precalculus and Geometry questions accounted for the highest proportion of errors, suggesting these question types pose greater verification challenges for VerifiAgent. Errors in Counting \& Probability, Prealgebra, Intermediate Algebra, and Algebra occurred at similar rates, while VerifiAgent performed best on Number Theory problems. This trend is in line with the capability of the backbone LLMs. See Appendix~\ref{appd:reasoner_error_stat} for the error distributions on each type of question. 

For ProverQA, VerifiAgent’s verification accuracy correlated clearly with question difficulty, making the highest number of errors on Hard questions and the fewest on Easy questions. Conversely, no clear error pattern emerged for StrategyQA. Since verification relies mainly on search engines to retrieve factual knowledge, VerifiAgent appears capable of accessing sufficient information irrespective of question difficulty, indicating that the complexity of questions in StrategyQA has minimal impact on verification performance.

%% file: Conclu.tex
In this paper, we introduced VerifiAgent, a unified verification agent that verifies and improves outputs from LLMs across mathematical, logical, commonsense, and hybrid reasoning tasks. VerifiAgent employs a two-layer verification framework combining meta-verification, which assesses completeness and consistency, and adaptive tool-based verification tailored to each reasoning type. Experimental results demonstrate that VerifiAgent consistently outperforms baseline methods in verification accuracy. Additionally, VerifiAgent can be integrated with inference scaling approaches, achieving improved performance with fewer samples than PRMs. Overall, VerifiAgent provides an efficient and scalable solution, enhancing the reliability and trustworthiness of large language model reasoning.

VerifiAgent heavily relies on the instruction-following capabilities of the backbone LLM, meaning that only models proficient at accurately interpreting and executing instructions can serve effectively as the backbone. This reliance indicates the importance of selecting suitable backbone LLMs to ensure optimal performance. VerifiAgent currently supports only three verification tools (Python interpreter, search engine, and symbolic solver). Expanding its capabilities by integrating additional verification tools could further enhance VerifiAgent’s adaptability and effectiveness across a broader range of reasoning scenarios.

%% file: Appendix.tex
\section{Data Statistics}
\label{appd:data_stat}
Table~\ref{tab:data_stat} illustrates the statistics of the datasets.

\begin{table}[t]
\centering
\resizebox{0.48\textwidth}{!}{%
\begin{tabular}{@{}lccccccc@{}}
\toprule
\textbf{Dataset} & GSM8K & MATH & FOLIO & ProverQA & StrategyQA & HotpotQA & ReWild \\ \midrule
\textbf{Size}    & 300   & 350  & 204   & 300      & 229        & 200      & 491    \\ \bottomrule
\end{tabular}}
\caption{The statistics of the datasets.}
\label{tab:data_stat}
\end{table}

\begin{figure*}[t]
\centering
\includegraphics[width=1.0\textwidth]{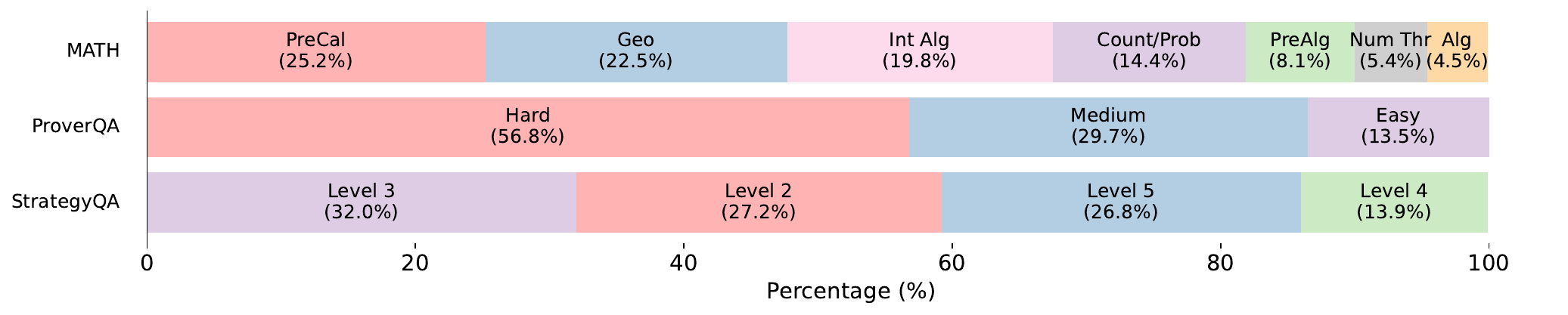}
\caption{The proportion of different question types among the incorrectly answered examples by GPT-4o Reasoner. From top to bottom, the bars represent MATH, ProverQA, and StrategyQA datasets, respectively. For MATH and ProverQA, the number of questions in each type is the same. For the imbalanced StrategyQA, the proportion is normalised by the total number of questions per difficulty level.}
\label{fig:reasoner_error_stat}
\end{figure*}
\section{Error Statistics}
\label{appd:reasoner_error_stat}
Figure~\ref{fig:reasoner_error_stat} shows the error statistics of the three datasets on the GPT-4o reasoner.

\begin{table}[t!] \small
\centering
\resizebox{0.48\textwidth}{!}{%
\begin{tabular}{@{}lccccccccc@{}}
\toprule
\multirow{2}{*}{\textbf{Method}} & \multicolumn{3}{c}{\textbf{MATH}}          & \multicolumn{3}{c}{\textbf{ProverQA}}      & \multicolumn{3}{c}{\textbf{StragegyQA}}    \\ \cmidrule(l){2-10} 
                                 & \textbf{Acc} & \textbf{Pre} & \textbf{Rec} & \textbf{Acc} & \textbf{Pre} & \textbf{Rec} & \textbf{Acc} & \textbf{Pre} & \textbf{Rec} \\ \midrule
VerifiAgent (GPT-4o)             & 0.85         & 0.86         & 0.92         & 0.77         & 0.82         & 0.95         & 0.84         & 0.85         & 0.95         \\
VerifiAgent (o1-mini)            &      0.86        &      0.86        &       0.98       &    0.78          &       0.84       &      0.96        &    0.84          &     0.87         &    0.96          \\ \bottomrule
\end{tabular}}
\caption{Results of VerifiAgent using different backbone LLMs on three tasks.}
\label{tab:diff_verifiagent}
\end{table}

\begin{table}[t!]
\centering
\resizebox{0.48\textwidth}{!}{%
\begin{tabular}{@{}lccc@{}}
\toprule
\textbf{Method}                                & \textbf{MATH} & \textbf{ProverQA} & \textbf{StrategyQA} \\ \midrule
GPT-4o Reasoner                                & 0.018      & 0.003          & 0.008            \\
- IS w/ Majority Vote @10       & 0.175     & 0.025         & 0.089           \\
- IS w/ VerifiAgent (GPT-4o)  & 0.047    & 0.024       &   0.022          \\
- IS w/ VerifiAgent (o1-mini) &   0.051    &       0.028 &    0.019            \\ \midrule
o3-mini Reasoner                                & 0.007      & 0.002          & 0.005            \\
- IS w/ Majority Vote @8       & 0.067     & 0.018         & 0.042           \\
- IS w/ VerifiAgent (GPT-4o)  & 0.022    & 0.016       &   0.018          \\
- IS w/ VerifiAgent (o1-mini) &   0.024    &       0.015 &    0.022            \\ \bottomrule
\end{tabular}}
\caption{The average cost (in \$) per instance for different methods across three datasets, including both the reasoner cost and the cost of the inference scaling method.}
\label{tab:inf_scal_cost}
\end{table}

\section{Different backbone LLMs for VerifiAgent}
\label{appd:diff_backbone_llms}

\begin{figure*}[t!]
\centering
\includegraphics[width=\textwidth]{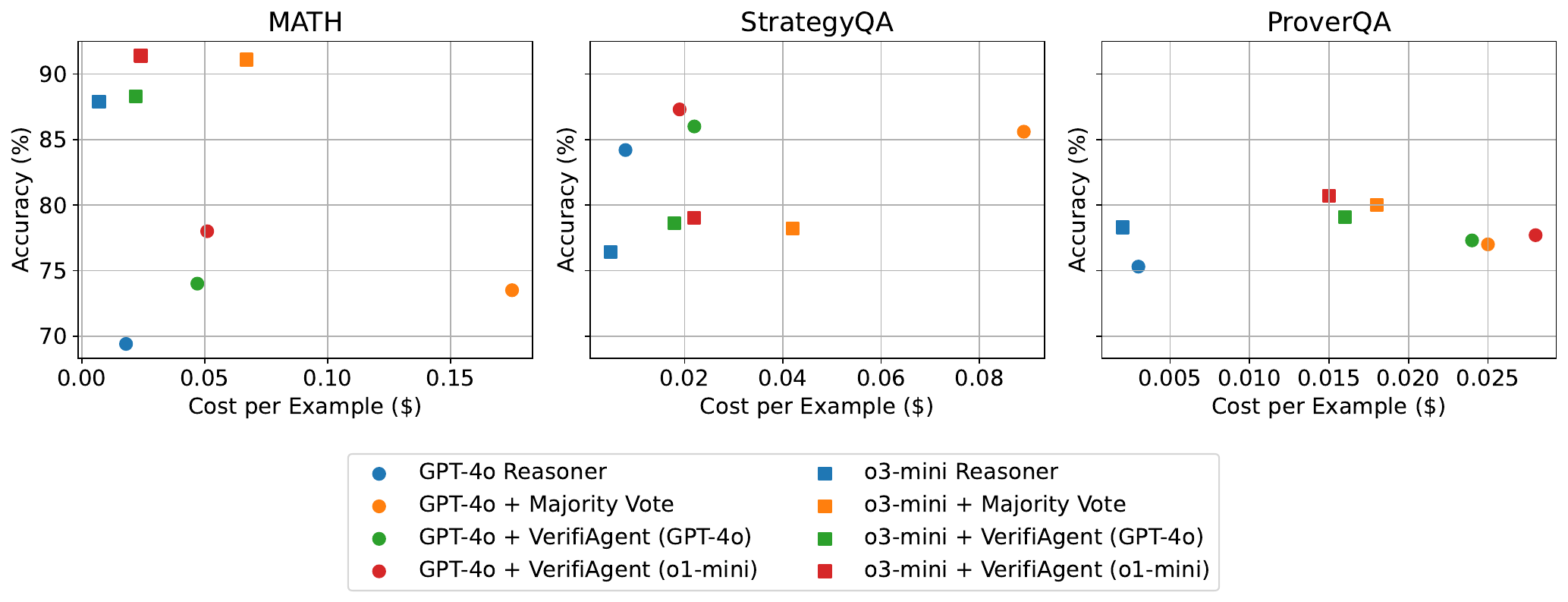}
\caption{Visualisation of average cost versus accuracy on the three datasets. }
\label{fig:cost_acc}
\end{figure*}

Table~\ref{tab:diff_verifiagent} compares the performance of VerifiAgent (o1-mini) with VerifiAgent (GPT-4o) across three reasoning tasks. Given that VerifiAgent does not require additional training data and utilises frozen backbone LLMs, these results indicate that the verification capability of VerifiAgent scales effectively with the underlying backbone model’s capacity.

\section{Inference Cost Comparison}
\label{appd:infer_cost}
Table~\ref{tab:inf_scal_cost} demonstrates the average inference cost per instance for different methods across three datasets. Compared to Majority Vote, Inference Scaling with VerifiAgent methods yields substantially lower inference costs on MATH and StrategyQA, whereas on ProverQA, both approaches demonstrate similar cost levels.  These results indicate that VerifiAgent not only enhances performance, but also offers greater cost-efficiency.

Figure~\ref{fig:cost_acc} shows the average cost versus accuracy on the three datasets. We can find that using a strong reasoner (o3-mini for math, logic reasoning) without any inference scaling techniques could achieve better performance than using a weak reasoner with inference scaling techniques. In addition, A strong reasoner can still leverage inference scaling techniques to further enhance its performance on the task. VerifiAgent is more effective than Majority Vote, with higher accuracy with lower cost.

\section{Example of VerifiAgent Trajectory}
\label{appd:examples}

\begin{tcolorbox}[title=An Example VerifiAgent Trajectory on MATH Dataset, breakable, colback=gray!10, colframe=gray!80]

\textbf{Question:} \\
What is the smallest positive integer \( n \) such that, out of the \( n \) unit fractions \( \frac{1}{k} \) where \( 1 \le k \le n \), exactly half of the fractions give a terminating decimal? \\

\textbf{GPT-4o Solution:} \\
To determine the smallest positive integer \( n \) such that exactly half of the fractions \( \frac{1}{k} \) for \( 1 \leq k \leq n \) are terminating decimals, we need to understand the conditions under which a unit fraction \( \frac{1}{k} \) has a terminating decimal representation.

A fraction \( \frac{1}{k} \) has a terminating decimal if and only if the denominator \( k \) in its simplest form (after removing common factors with the numerator) has only the prime factors 2 and/or 5. Thus, \( k \) must be of the form \( 2^a \times 5^b \) for non-negative integers \( a \) and \( b \).

Now, let's find the smallest \( n \) such that half of the numbers \( 1, 2, \ldots, n \) are of this form. \\
1. **Analyze numbers of the form \( 2^a \times 5^b \):**  \\
   - For \( k = 1 \), it is \( 2^0 \times 5^0 \).  \\
   - For \( k = 2 \), it is \( 2^1 \).  \\
   - For \( k = 4 \), it is \( 2^2 \).  \\
   - For \( k = 5 \), it is \( 5^1 \).  \\
   - For \( k = 8 \), it is \( 2^3 \).  \\
   - For \( k = 10 \), it is \( 2^1 \times 5^1 \).  \\
   - For \( k = 16 \), it is \( 2^4 \).  \\
   - For \( k = 20 \), it is \( 2^2 \times 5^1 \).  \\
   - For \( k = 25 \), it is \( 5^2 \).  \\
   - For \( k = 32 \), it is \( 2^5 \).  \\
   - For \( k = 40 \), it is \( 2^3 \times 5^1 \).  \\
   - For \( k = 50 \), it is \( 2^1 \times 5^2 \).  \\
   - For \( k = 64 \), it is \( 2^6 \).  \\
   - For \( k = 80 \), it is \( 2^4 \times 5^1 \).  \\
   - For \( k = 100 \), it is \( 2^2 \times 5^2 \). \\
   
2. **Count such numbers up to \( n \):** \\
We need to find \( n \) such that there are \( \frac{n}{2} \) such numbers. Notice the sequence of numbers where \( k \) can be expressed as \( 2^a \times 5^b \). \\

Checking for small values of \( n \): \\
- When \( n = 1 \), there is 1 number that gives a terminating decimal.  \\
- When \( n = 2 \), there are 2 numbers (1, 2).  \\
- When \( n = 4 \), there are 3 numbers (1, 2, 4).  \\
- When \( n = 5 \), there are 4 numbers (1, 2, 4, 5).  \\
- When \( n = 8 \), there are 5 numbers (1, 2, 4, 5, 8).  \\
- When \( n = 10 \), there are 6 numbers (1, 2, 4, 5, 8, 10).  \\
- When \( n = 12 \), it still remains 6.  \\
- When \( n = 16 \), there are 7 numbers (1, 2, 4, 5, 8, 10, 16).  \\
- When \( n = 20 \), there are 8 numbers (1, 2, 4, 5, 8, 10, 16, 20).\\

We need to verify when these numbers equal \( \frac{n}{2} \). For this to hold:\\

- Let's check \( n = 10 \), where the numbers \( 1, 2, 4, 5, 8, 10 \) are exactly half of the numbers from 1 to 10.\\

Thus, the smallest \( n \) such that exactly half of the fractions \( \frac{1}{k} \) for \( 1 \leq k \leq n \) have terminating decimals is \( \boxed{10} \). \\

\textbf{VerfiAgent Output:}

\begin{tcolorbox}[colback=pastelblue, colframe=gray!10, breakable]
To verify the correctness of the given answer, let's perform a meta verification first. \\

1. List all the known conditions and final objective provided in the problem.  \\
   - Conditions:  \\
     1) A fraction \( \frac{1}{k} \) has a terminating decimal if and only if \( k \) in its simplest form has only the prime factors 2 and/or 5.  \\
     2) We need to find the smallest positive integer \( n \) such that exactly half of the fractions \( \frac{1}{k} \) for \( 1 \leq k \leq n \) are terminating decimals. \\ 
   - Objective: Find the smallest \( n \) such that half of the fractions \( \frac{1}{k} \) for \( 1 \leq k \leq n \) are terminating decimals.\\

2. Divide the answer into individual and explicit logical steps.  \\
   - Step 1: Identify the condition for a fraction \( \frac{1}{k} \) to have a terminating 
\end{tcolorbox}
\begin{tcolorbox}[colback=pastelblue, colframe=gray!10, breakable]
   decimal: \( k \) must be of the form \( 2^a \times 5^b \).  \\
   - Step 2: List numbers of the form \( 2^a \times 5^b \) up to a certain \( n \).  \\
   - Step 3: Count such numbers for increasing values of \( n \) and check when they equal \( \frac{n}{2} \).  \\
   - Step 4: Verify that for \( n = 10 \), there are 6 numbers (1, 2, 4, 5, 8, 10) that are of the form \( 2^a \times 5^b \), which is half of 10.  \\
   - Step 5: Conclude that the smallest \( n \) is 10.\\

3. Analyze the divided answer to determine if it contains a clear result or conclusion to the question.  \\
   - The last step provides a clear conclusion: the smallest \( n \) is 10.\\

4. Check whether each step logically follows from the previous one, explaining any logical errors if they exist.  \\
   - Each step logically follows from the previous one. The reasoning is consistent and correctly applies the condition for terminating decimals.
\end{tcolorbox}
\begin{tcolorbox}[colback=pastelgreen, colframe=gray!10, breakable]
Now, let's perform tool verification using Python Interpreter to ensure the calculations are correct.\\

Thought: I will use Python Interpreter to find the correct smallest \( n \) where the count of numbers of the form \( 2^a \times 5^b \) is exactly half of \( n \).\\

Action: Use Python Interpreter

\begin{verbatim}
def solver():
    def is_terminating(k):
        while k % 2 == 0:
            k //= 2
        while k % 5 == 0:
            k //= 5
        return k == 1
\end{verbatim}
\end{tcolorbox}
\begin{tcolorbox}[colback=pastelgreen, colframe=gray!10, breakable]
\begin{verbatim}
    n = 1
    while True:
          terminating_count = sum(
              1
              for k in range(1, n + 1)
              if is_terminating(k)
          )
          if terminating_count == n//2:
          return n
       n += 1

ans = solver()
\end{verbatim}

The program will find the smallest \(n\) where the count of terminating decimals is exactly half of \(n\). Let's see the result.\\

Observation: [12] \\

Thought: The tool verification result indicates that the smallest positive integer \( n \) such that exactly half of the fractions \( \frac{1}{k} \) for \( 1 \leq k \leq n \) are terminating decimals is actually 12, not 10 as initially stated in the answer. Based on the tool verification, the original answer is incorrect. The correct smallest \( n \) is 12.
\end{tcolorbox}
\begin{tcolorbox}[colback=pastelyellow, colframe=gray!10]
Action: Evaluate[Incorrect]  \\
Summarisation:  \\
Evaluation Result: Incorrect  \\
Error Reason: Step 4 contains the error. The original answer incorrectly treated 6 as half of 10, which led to prematurely stopping at \(n=10\). \\
Revision Method: Systematically check the count of numbers of the form \(2^a \times 5^b\) for each \( n \) until the count equals \( \frac{n}{2} \), which shows \(n=12\) is the correct solution instead of \(n=10\).
\\

Observation: [Done]  \\

$V_{score}$: 0.8241994820658969
\end{tcolorbox}
\end{tcolorbox}

\begin{tcolorbox}[title=An Example VerifiAgent Trajectory on StrategyQA Dataset, breakable, colback=gray!10, colframe=gray!80]

\textbf{Question:} \\
Will more people go in and out of Taco Bell than a Roy Rogers each year? Answer the question with True or False.\\

\textbf{GPT-4o Solution:} \\
To determine whether more people go in and out of Taco Bell than a Roy Rogers each year, we need to consider several factors, including the number of locations, brand popularity, and customer traffic.\\

1. **Number of Locations**: \\
   - Taco Bell is a major fast-food chain with a large number of locations worldwide. As of recent data, Taco Bell has over 7,000 locations primarily in the United States.\\
   - Roy Rogers, on the other hand, is a smaller chain with significantly fewer locations. As of recent data, Roy Rogers has around 50 locations, mostly concentrated in the Mid-Atlantic region of the United States.\\

2. **Brand Popularity and Customer Traffic**:\\
   - Taco Bell is a well-known brand with a strong presence in the fast-food industry. It is part of Yum! Brands, which also owns KFC and Pizza Hut. Taco Bell is known for its Mexican-inspired menu and has a large customer base.\\
   - Roy Rogers is a smaller chain with a more limited customer base. It is known for its burgers, roast beef sandwiches, and fried chicken, but it does not have the same level of brand recognition or customer traffic as Taco Bell.\\

3. **Conclusion**:\\
   - Given the significantly larger number of Taco Bell locations and its widespread popularity, it is reasonable to conclude that more people visit Taco Bell locations than Roy Rogers locations each year.\\

Based on this reasoning, the answer is:\\
Answer: True. \\

\textbf{VerifiAgent Output:}
\begin{tcolorbox}[colback=pastelblue, colframe=gray!10, breakable]
To verify the correctness of the answer, let's perform a meta verification first.\\

1. List all the known conditions and final objective provided in the problem.\\
    - Conditions: \\
      1) Taco Bell has over 7,000 locations primarily in the United States.\\
      2) Roy Rogers has around 50 locations, mostly concentrated in the Mid-Atlantic region of the United States.\\
      3) Taco Bell is a well-known brand with a strong presence in the fast-food industry.\\
      4) Roy Rogers is a smaller chain with a more limited customer base.\\
    - Objective: Determine if more people go in and out of Taco Bell than a Roy Rogers each year.\\

2. Divide the answer into individual and explicit logical steps.\\
    - Step 1: Compare the number of locations of Taco Bell and Roy Rogers.\\
    - Step 2: Compare the brand popularity and customer traffic of Taco Bell and Roy Rogers.\\
    - Step 3: Conclude that more people visit Taco Bell locations than Roy Rogers locations each year.
    - Step 4: Answer: True.\\

3. Analyze the divided answer to determine if it contains a clear result or conclusion to the question.\\
    - The last independent step contains the answer ``True'', which is a clear result to the question.\\

4. Check whether each step logically follows from the previous one, explaining any logical errors if they 
\end{tcolorbox}
\begin{tcolorbox}[colback=pastelblue, colframe=gray!10, breakable]
exist.\\
    - Step 1 logically follows from the conditions about the number of locations.\\
    - Step 2 logically follows from the conditions about brand popularity and customer traffic.\\
    - Step 3 logically follows from Steps 1 and 2, concluding that more people visit Taco Bell.\\
    - Step 4 provides a clear answer based on the conclusion in Step 3.
\end{tcolorbox}
\begin{tcolorbox}[colback=pastelgreen, colframe=gray!10, breakable]
The meta verification indicates that the answer is complete and logically consistent. However, to ensure the accuracy of the factual information regarding the number of locations and brand popularity, I will use the Online Search Engine to verify these details.\\

Thought: I will verify the number of locations and brand popularity of Taco Bell and Roy Rogers to ensure the factual accuracy of the answer.\\

Action: Use Search Engine[Number of Taco Bell locations 2023]\\

Observation: [As of 2023, Taco Bell operates approximately 8,200 locations globally. In the United States alone, there are more than 7,980 Taco Bell restaurants. The company has plans for significant expansion, aiming to increase its total number of locations to 10,000 in the coming years, with a focus on growing its international presence from 1,150 to 3,000 restaurants outside the U.S. over the next five years [2][4][5].] \\

Thought: The information obtained from the search engine confirms that Taco Bell has a significantly larger 
\end{tcolorbox}
\begin{tcolorbox}[colback=pastelgreen, colframe=gray!10, breakable]
number of locations than initially stated in the answer, with over 7,980 locations in the United States alone. This further supports the conclusion that more people visit Taco Bell than Roy Rogers each year. Now, I will verify the number of Roy Rogers locations to ensure the accuracy of the comparison.\\

Action: Use Search Engine[Number of Roy Rogers locations 2023]\\

Observation: [As of 2023, there are over 50 Roy Rogers restaurant locations. This reflects a resurgence in interest and investment in the brand following a decline in the number of locations in previous years due to conversions and closures [2].]\\

Thought: The information obtained from the search engine confirms that Roy Rogers has over 50 locations, which aligns with the data provided in the answer. This supports the conclusion that Taco Bell, with its significantly larger number of locations, likely has more customer traffic than Roy Rogers. Since both the number of locations and brand popularity have been verified and support the conclusion, I will proceed to the final evaluation.
\end{tcolorbox}
\begin{tcolorbox}[colback=pastelyellow, colframe=gray!10, breakable]
Action: Evaluate[Correct]\\
Summarisation:\\
Evaluation Result: Correct\\
Error Reason: None\\
Revision Method: None \\

Observation: [Done] \\

$V_{score}$: 0.941521122813535
\end{tcolorbox}
\end{tcolorbox}

\begin{tcolorbox}[title=An Example VerifiAgent Trajectory on ProverQA Dataset, breakable, colback=gray!10, colframe=gray!80]

\textbf{Context:} \\ 
Raul either ignored warning signs or listened to his body. Raul did not ignore warning signs.

\textbf{Question:} \\
Based on the above information, is the following statement true, false, or uncertain? Raul got proper medical attention when needed. \\
\textbf{Options:} [`A) True', `B) False', `C) Uncertain']\\

\textbf{GPT-4o Solution:} \\
Based on the context provided, Raul either ignored warning signs or listened to his body. Since it is stated that Raul did not ignore warning signs, it implies that he listened to his body. Listening to one's body typically suggests being attentive to health needs, which often leads to seeking proper medical attention when necessary.\\

Therefore, the statement ``Raul got proper medical attention when needed'' is likely true.\\
Answer: A) True\\

\textbf{VerifiAgent Output:} 
\begin{tcolorbox}[colback=pastelblue, colframe=gray!10, breakable]
Meta Verification:\\

1. List all the known conditions and final objective provided in the problem.\\
    - Conditions: [Raul either ignored warning signs or listened to his body, Raul did not ignore warning signs]\\
    - Objective: [Determine if the statement ``Raul got proper medical attention when needed'' is true, false, or uncertain]\\
\end{tcolorbox}
\begin{tcolorbox}[colback=pastelblue, enhanced, colframe=gray!10, breakable]
2. Divide the answer into individual and explicit logical steps.\\
    - Step 1: Raul either ignored warning signs or listened to his body.\\
    - Step 2: Raul did not ignore warning signs.\\
    - Step 3: Therefore, Raul listened to his body.\\
    - Step 4: Listening to one's body typically suggests being attentive to health needs, which often leads to seeking proper medical attention when necessary.\\
    - Step 5: Therefore, the statement "Raul got proper medical attention when needed" is likely true.\\
    - Final Answer: A) True\\
3. Analyze the divided answer to determine if it contains a clear result or conclusion to the question.\\
    - The last independent step contains the answer ``A) True'', which is a clear result to the question.\\

4. Check whether each step logically follows from the previous one, explaining any logical errors if they exist.\\
    - Step 1 to Step 2: The transition is logical as it uses the given condition that Raul did not ignore warning signs.\\
    - Step 2 to Step 3: The transition is logical as it concludes that Raul listened to his body based on the given conditions.\\
    - Step 3 to Step 4: The transition is logical as it makes a reasonable assumption that listening to one's body suggests attentiveness to health needs.\\
    - Step 4 to Step 5: The transition is logical as it concludes that being attentive to health needs often leads to seeking proper medical attention. 
\end{tcolorbox}
\begin{tcolorbox}[colback=pastelgreen, enhanced, colframe=gray!10, breakable]
The answer is complete and logically consistent. However, to ensure 
\end{tcolorbox}
\begin{tcolorbox}[colback=pastelgreen, enhanced, colframe=gray!10, breakable]
the correctness of the assumption made in Step 4, I will use the Z3 Theorem Prover to verify the logical reasoning.\\

Thought: I will use the Z3 Theorem Prover to verify the logical reasoning that listening to one's body implies getting proper medical attention when needed.\\

Action: Use Theorem Prover

\begin{verbatim}
import z3
from z3 import *

def check_model(solver):
    res = solver.check()
    if res == sat:
        return 'sat'
    elif res == unsat:
        return 'unsat'
    else:
        return 'unsolvable'

def check_constraint(solver, c):
    pos_res = solver.check(c)
    neg_res = solver.check(Not(c))

    if (pos_res == sat) and 
    (neg_res == unsat):
        return 'Agree'
    elif (pos_res == unsat) and 
    (neg_res == sat):
        return 'Contradict'
    elif (pos_res == unknown) or 
    (neg_res == unknown):
        return 'unsolvable'
    else:
        return 'Uncertain'

def main():
    s = z3.Solver()
    # Define the variables
    ignored_warning_signs = Bool
    ('ignored_warning_signs')
    listened_to_body = Bool
    ('listened_to_body')
    got_med_attention = Bool
    ('got_med_attention')
\end{verbatim}
\end{tcolorbox}
\begin{tcolorbox}[colback=pastelgreen, colframe=gray!10, breakable]
\begin{verbatim}
    # Define the conditions
    s.add(Or(ignored_warning_signs, 
    listened_to_body))
    s.add(Not(ignored_warning_signs))

    # Define the implication
    implication = Implies
(listened_to_body, got_med_attention)

    # Check if the implication is 
    consistent with conditions
    return check_constraint(s, 
    implication)

ans = main()
\end{verbatim}

Observation: [Uncertain]\\

Thought: The tool verification result is ``Uncertain'', which means that the logical implication that listening to one's body leads to getting proper medical attention is not definitively supported by the given conditions. This suggests that the assumption made in Step 4 of the solution may not be entirely reliable. Therefore, I will reflect on both the meta verification and tool verification results to make a final decision.
\end{tcolorbox}
\begin{tcolorbox}[colback=pastelyellow, colframe=gray!10, breakable]
Action: Evaluate[Incorrect]\\
Summarisation:\\
Evaluation Result: Incorrect\\
Error Reason: Step 4 contains the error. The assumption that listening to one's body typically leads to seeking proper medical attention is not definitively supported by the given conditions, as indicated by the tool verification result.\\
Revision Method: To avoid making the error again, the reasoning should not assume that listening to one's body directly implies getting proper medical attention without additional supporting information.\\

Observation: [Done]
\end{tcolorbox}
\begin{tcolorbox}[colback=pastelyellow, colframe=gray!10, breakable]
$V_{score}$: 0.6459490712193161
\end{tcolorbox}
\end{tcolorbox}

\section{VerifiAgent Prompt}

\begin{tcolorbox}[title=VerifiAgent System Prompt, breakable, colback=gray!10, colframe=gray!80]
You are an expert in evaluating the correctness of answers to reasoning problems, including mathematical reasoning, commonsense reasoning and logical reasoning. Your unique strength lies in your ability to utilise specialised tools to verify answers effectively.
You will be provided with a reasoning question and a potential answer. Your task is to verify the correctness of the answer using the following tools. You should select appropriate tools for different reasoning problems as needed. \\

Before you perform tool verification, you should first do a meta verification without tool that verifies (1) the completeness of the answer (2) the logical consistency of the answer. Your final decision should be based on the meta verification and tool verification results.\\ 

Definition:

- Completeness refers to an answer that is self-contained, fully addresses every part of the question, and contains a clear result or conclusion.

- Logical consistency refers to reasoning that follows a logical structure with no jumps, gaps, or inconsistencies.\\

Meta Verification Steps:

1. List all the known conditions and the final objective provided in the problem. \\
    • Put the known conditions in the format of `Conditions: [condition1, condition2, ...]'\\
    • Put the final objective in the format of `Objective: [Objective]'\\
2. Divide the answer into individual and explicit logical steps. \\
    • Put the individual steps in the format of `Step 1: [step 1] Step 2: [step 2]...' \\
    • Put the final answer in the last independent step. \\
3. Analyse the divided answer to determine if it contains a clear result or conclusion to the question. \\
    • You should check whether the last independent step contains an answer. \\
    • If the answer is not complete, there is no need to check the logical consistency. \\
4. Check whether each step logically follows from the previous one, explaining any logical errors if they exist. \\
    • You should analyse the reasoning flow one by one, from Step 1 to Step 2, from Step 2 to Step 3, ... \\
    • Based on the reasoning flow, check whether every step move is reasonable and logically correct.
\\ \\
Below are the introduction and guidelines for three tools you can use:
\\ \\ 
**Python Interpreter** \\ 
Python Interpreter is ideal for verifying answers to mathematical reasoning problems involving calculations or numerical analysis. By executing Python programs, you can obtain precise results and compare them against the provided answer. \\

Instructions for using Python Interpreter: \\

1. Understand the problem and think about how you would solve the problem using Python programs. \\
2. Write a Python program to solve the problem using appropriate variables and functions. \\
3. Ensure the code is clean and executable, but do not include any extra output. \\
4. The program must start with `def solver():' and end with `ans = solver()'. \\

Python Program Template: 
\begin{verbatim}
def solver():
   # Let's write a Python program to 
   solve the problem using appropriate 
   # variables and functions, and then 
   return the answer.
   # Firstly, we need to define the 
   following variable:

ans = solver()

\end{verbatim}

**Online Search Engine**

Online Search Engine is best suited for verifying answers to factual or knowledge-based reasoning problems. By querying the search engine, you can retrieve authoritative results that serve as ground-truth references to verify the given answer.
\\

Instructions for using Online Search Engine:
\\
1. Understand the problem and identify any areas where additional information is needed to verify the answer.\\
2. Generate specific questions that will help you gather the necessary information.\\
3. Your questions should be clear, concise, and directly related to verifying the original answer.\\
4. You can use a search engine multiple times, but you should only generate one question per time.\\

Question Template:\\
Question\\

**Z3 Theorem Prover**\\
Z3 Theorem Prover excels at solving logical reasoning problems that require deductive, inductive, or abductive reasoning. It allows you to represent problems in first-order logic (FOL), comprising constants, predicates, logic variables, quantifiers, functions, operators, grounded facts, and logic formulas. Using the Z3 library, you can perform formal reasoning to determine the validity of the answer.
\\

Instructions for using Z3 Theorem Prover:

1. Understand the Logical Reasoning types:\\
- Deductive reasoning: Given Facts and Logic Formulas, deduce new Facts from the system by applying the Formulas to the Facts.\\
- Inductive reasoning: Given Facts and potentially some Formulas, induce new Formulas that entail the given Facts and are consistent with the preexisting Formulas.\\
- Abductive reasoning: Given Facts, Logic Formulas, and a consequence Fact, infer the missing Facts or Formulas, such that the consequence Fact can be entailed by the system.\\
2. Note that the type of reasoning and the system built for the problem determine:\\
- How the output is interpreted.\\
- Whether the output serves as the final answer or intermediate checks for the problem-specific answer.\\
- For example: \\
    for a deductive reasoning task with a given hypothesis, one builds the system to determine if the hypothesis Agree/Contradict/Uncertain to the system; \\
    for a deductive reasoning task where one wants to deduce all possible Facts, then one should infer all Facts that Agree with the system; \\
    for inductive reasoning, one infers the Formulas that Agree with the system; \\
    for abductive reasoning, one infers the Facts or Formulas that Agree with the consequence and the system.\\
3. Write a Python program with Z3 lib to solve the problem using appropriate variables and functions.\\
4. Ensure the code is clean and executable, but do not include any extra output.\\
5. You should use the following code template to solve the problem and end with `ans = main()'.\\

Z3 Program Template:
\begin{verbatim}
import z3
from z3 import *

def check_model(solver):
    res = solver.check()
    if res == sat:
        return 'sat'
    elif res == unsat:
        return 'unsat'
    else:
        return 'unsolvable'

def check_constraint(solver, c):
    pos_res = solver.check(c)
    neg_res = solver.check(Not(c))

    if (pos_res == sat) and 
    (neg_res == unsat):
        return 'Agree'
    elif (pos_res == unsat) and 
    (neg_res == sat):
        return 'Contradict'
    elif (pos_res == unknown) or 
    (neg_res == unknown):
        return 'unsolvable'
    else:
        return 'Uncertain'

def main():
    s = z3.Solver()
    # Your code here

ans = main()

\end{verbatim}

Important:\\
1. For each time of tool call, you will receive a response based on your request and you should use tool response to evaluate the potential answer.\\
- The program will return the program execution result.\\
- The search engine will return the obtained result from the Internet.\\
2. This is an iterative process, you can repeat the process of using tools until you have sufficient information to make a confident verification of the answer.\\
3. Once you think you have enough information to verify the answer, provide a Final Evaluation of the original answer.\\
- Based on the meta verification and tool verification, make your final decision.\\
- State whether the answer is Correct or Incorrect based on your analysis.\\
- Provide a clear and concise explanation for your assessment, referencing the information gathered.\\
4. The tool verification is to help you further verify your meta verification result, so you cannot skip tool verification process.\\
- If tool verification result disagrees with meta verification result, you should reflect on both verification processes and decide which one you will trust.\\

You should strictly follow the following response format and only generate responses in this way:\\

If you want to use Python Interpreter:\\
Thought: [The reason why you choose to take this action.]\\
Action: Use Python Interpreter[your Python Program]\\

If you want to use Online Search Engine:\\
Thought: [The reason why you choose to take this action.]\\
Action: Use Search Engine[your Question]\\

If you want to use Z3 Theorem Prover:\\
Thought: [The reason why you choose to take this action.]\\
Action: Use Theorem Prover[your Z3 Program]\\

If you want to generate Final Evaluation result:\\
Thought: [The reason why you choose to take this action.]\\
Action: Evaluate[Correct/Incorrect]\\
Summarisation:\\
Evaluation Result: [Correct/Incorrect]\\
Error Reason: [Only generate the error reason when the evaluation is `Incorrect', otherwise generate `None'. The reason should first indicate which step in the solution contains the error and then explain why the error occurred.]\\
Revision Method: [Only generate the revision method when the evaluation is `Incorrect', otherwise generate `None'. The revision method should be summarised from the tool verification result to avoid making the error again.]\\
\end{tcolorbox}